\documentclass[10pt,twocolumn]{article}

\pdfoutput=1

\usepackage{cvpr}
\usepackage{times}
\usepackage{epsfig}
\usepackage{graphicx}
\usepackage{amsmath}
\usepackage{amssymb}
\usepackage{algorithm}
\usepackage{algorithmic}
\usepackage{accents}

\newcommand{\DeltaOverEq}{\accentset{\Delta}{=}}

\newtheorem{theorem}{Theorem}[section]
\newtheorem{lemma}[theorem]{Lemma}

\newenvironment{proof}[1][Proof]{\begin{trivlist}
\item[\hskip \labelsep {\bfseries #1}]}{\end{trivlist}}

\newcommand{\qed}{\nobreak \ifvmode \relax \else
      \ifdim\lastskip<1.5em \hskip-\lastskip
      \hskip1.5em plus0em minus0.5em \fi \nobreak
      \vrule height0.75em width0.5em depth0.25em\fi}

\cvprfinalcopy 

\def\cvprPaperID{1202} 

\ifcvprfinal\fi
\begin{document}

\title{Convolutional Tables Ensemble: classification in microseconds}

\author{Aharon Bar-Hillel\\
Microsoft Research\\
Haifa\\
{\tt\small aharonb@microsoft.com}
\and
Eyal Krupka\\
Microsoft Research\\
Haifa\\
{\tt\small eyalk@microsoft.com}
\and
Noam Bloom\\
Microsoft Research\\
Haifa\\
{\tt\small t-noblo@microsoft.com}
}

\maketitle

\begin{abstract}

We study classifiers operating under severe classification time constraints, corresponding to $1-1000$ CPU microseconds, using Convolutional Tables Ensemble (CTE), an inherently fast architecture for object category recognition. The architecture is based on convolutionally-applied sparse feature extraction, using trees or ferns, and a linear voting layer. Several structure and optimization variants are considered, including novel decision functions, tree learning algorithm, and distillation from CNN to CTE architecture. Accuracy improvements of~$24-45\%$ over related art of similar speed are demonstrated on standard object recognition benchmarks. Using Pareto speed-accuracy curves, we show that CTE can provide better accuracy than Convolutional Neural Networks (CNN) for a certain range of classification time constraints, or alternatively provide similar error rates with $5-200\times$ speedup.

\end{abstract}

\section{Introduction}
\label{sec:Intro}

Practical object recognition problems often have to be solved under severe computation and time constraints. 
Some examples of interest are natural user interfaces, automotive active safety, robotic vision or sensing for the Internet of Things (IoT). Often the problem is to obtain high accuracy in real time, on a low power platform, or in a background process that can only utilize a small fraction of the CPU. In other cases the classifier is part of a cascade, or a complex multiple-classifier system. The accuracy-speed trade-off has thus been widely discussed in the literature, and various architectures have been suggested~\cite{He2013,Lebedev2014,DPM_LSH13,Benenson2012,AFS2013,ShottonSKFFBCM13}. Here we focus on the extreme end of this trade-off and ask: how accurate can we get for classifiers working in $1-1000$ CPU microseconds.

As a thought experiment, the fastest classifier possible would be the one concatenating all the pixel values into a single index, and then using this index to access a table listing the labels of all possible images. Of course, this is not feasible due to the exponential requirements of memory and training set size, but this limit case points us the direction to follow. The actual architecture we pursue compromises this limit idea in two main ways: 
first, instead of encoding the whole image with a single index, the image is treated as a dense set of patches where each patch is encoded using the same parameters. This is analogous to a convolutional layer in a convolutional neural network (CNN)~\cite{MNist1998}, where the same set of filters is applied at each image position. Second, instead of describing a patch using a single long index, it is encoded with a set of short indices, that are used to access a set of reasonable-size tables. Votes of all tables at all positions are combined linearly to yield the classifier outputs. Variants of this architecture have been used successfully mainly for classification of depth images~\cite{DFE, ShottonSKFFBCM13}. Here we explore this regime for visual recognition in general, under the term Convolutional Tables Ensemble (CTE). 


The idea of applying the same feature extraction on a dense locations grid is very old and influential in vision, and is a key tenet in CNNs, the state-of-the-art in object recognition. 
It provides a good structural prior in the form of translation invariance. Another advantage lies in enhanced sample size for learning local feature parameters, since these can be trained from (number of training images)$\times$(number of image patches) instances. 
The architectures we consider here are not deep in the CNN sense, and correspond to a single convolutional layer, followed by spatial pooling.

The main vessel we use for obtaining high classification speed is the utilization of table-based feature extractors, instead of heavier computations such as applying a large set of filters in a convolutional layer. In table-based feature extraction, the patch is characterized using a set of fast bit functions, such as a comparison between two pixels.
$K$ bits are extracted and concatenated into a word. 
This word is then used as an index into a set of weight tables, one per class, and the weights extracted provide the classes support from this word.
Support weights are accumulated across many tables and all image positions, and the label is decided according to 
the highest scoring class.

The power of this architecture is in the combination of fast-but-rich features with a high capacity classifier. Using $K$ quick bit functions, the representation considers all their $2^K$ combinations as features.
The representation is highly sparse, with $\frac{1}{2^K}$  of the features active at each position. The classifier is linear, but it operates over numerous highly non linear features. 
For $M$ tables and $C$ classes, the number of weights to optimize is $M2^KC$, which can be very high even for modest values of $M,K,C$. The architecture hence requires a large training set to be used, and it effectively trades training sample size for speed and accuracy.

Pushing the speed-accuracy envelope using this architecture requires making careful structural and algorithmic choices. First, bit functions and image preprocessing should be chosen. We start with the simple functions employed in~\cite{DFE, ShottonSKFFBCM13}, which were suitable for depth images, and extend them using gradient and color based channels and features employed in~\cite{Dollar2009}. Another type of bit function introduced are spatial bits stating the rough location of the patch, which enable us to combine global and local pooling. 
A second important choice is between conditional computation of bit functions, leading to tree structures like used in~\cite{ShottonSKFFBCM13}, and unconditional computation as in fern structures~\cite{DFE}. 
While trees 
may enable higher accuracy, ferns are better suited for vector processing (such as SSE instructions) and thus provide significant speed advantages. We explore between these ends empirically using a 'long tree' structure, whose configuration enables testing 
intermediate structures.

Several works have addressed the challenges of learning a tables-based classifier~\cite{Geurts2006, Bosch07, Tola08, Lepetit2006, RF_Cambridge, DFE, Ren_2014_CVPR}. These vary 
in optimization effort from extremely random forests~\cite{Geurts2006} to global optimization of table weights and greedy forward choice of bit functions~\cite{Ren_2014_CVPR,DFE}. Our approach builds on previous approaches, mostly~\cite{DFE}, and extends them with new possibilities. We learn the table ensemble by adding one table at a time, using a framework similar to the 'anyboost' algorithm~\cite{Mason2000,BarHillel2005}. Training iterates between minimizing a global convex loss, differentiating this loss w.r.t. examples, and using these gradients to guide construction of the next table. For the global optimization we used two main options: an SVM loss as used in~\cite{DFE}
and a softmax loss as commonly used in CNN training. For the optimization of the bit function parameters in a new fern/tree we developed
several options: forward bit selection, iterative bit replacement, and iterative local refinement. In some cases, such as the threshold parameters of certain bits, an algorithm providing the optimal solution is suggested. The algorithms considered are described in Section~\ref{sec:Method}. 

Since CTEs can be much faster than CNNs, while the latter excel at accuracy, one would naturally like to merge their advantages if possible. 
In several recent studies~\cite{Ba2014, Hinton2015,Romero2015}, the output of an accurate but computationally expensive classifier is used to train another classifier, with a different and often computationally cheaper architecture. We made a preliminary attempt to use this technique, termed distillation in~\cite{Hinton2015}, to train a CTE classifier with a CNN teacher, with encouraging results on the MNIST data. 

In Section~\ref{sub:Ablations} we present experiments demonstrating the performance gains of our techniques by comparison with the DFE method of~\cite{DFE}, ablation studies, fern-tree trade-off experiments, and distillation results. We use 
several publicly available object recognition benchmarks: MNIST~\cite{MNist1998}, CIFAR-10~\cite{Krizhevsky2009}, SVHN~\cite{Netzer2011} and 3-HANDPOSE~\cite{DFE}. CTE achieves error improvements of $24-45\%$ over~\cite{DFE}, with $38\%$ improvement on 3-HANDPOSE, the original data used in~\cite{DFE}. For MNIST, we were able to train a CTE with $0.45\%$ error and close to $100\mu s$ running time using the distillation technique. Even higher accuracy of $0.39\%$ error can be obtained with a tree-based CTE, with some cost in running time. 

In section~\ref{sub:AS} we systematically experimented with CTE configurations to obtain accuracy-speed trade-off graphs for the datasets mentioned. These graphs are compared to similar graphs obtained for CNNs. For the latter we trained NIN networks~\cite{Lin2013_NIN}, combining state-of-the-art accuracy with significant speed advantages, and further accelerated them by scaling parameters of breadth, NIN output dimension and convolution stride. Our results indicate that for a highly restricted CPU budget CTEs provide significantly better accuracy than CNNs, or conversely, CTEs can obtain the same error with a CPU budget lower by $5-200X$. Typically this is true for classifiers operating below $100$ microseconds on a single CPU thread. For the very low CPU bound domain CTE can still provide useful results, whereas CNNs completely break. 
This makes CTEs a natural architecture choice for that domain. Alternatives to CTE may be provided by the literature dealing with CNN acceleration~\cite{Lebedev2014,Vanhoucke2011, Liu_2015_CVPR}. 
However obtaining the speed gains made possible by CTEs using such techniques is far from trivial.


In summary, our main contribution in this paper is two-fold: First, we develop new algorithms in the CTE framework, improving upon related similar art and extending the framework to general object recognition. Second, we pose an alternative to CNN which enables improved accuracy at the highly CPU constrained regime. Short concluding remarks are given in Section~\ref{sec:Conclusions}.


\section{Convolutional Tables Ensemble}
\label{sec:Method}

\begin{algorithm}[t]
\caption{\textbf{\label{alg:CTE_test}}Convolutional Tables Ensemble: Classification}

{

\textbf{\small{{Input}}}{\small{{: An image I of size $S_{x}\times S_{y} \times D $,}}}{\small \par}

{\small{{ classifier parameters $(\Theta^{m},A^{m},W^{m,c},T^c)_{m=1,c=1}^{M,C}$ }}}{\small \par}

{\small{$A^{m}\subset\{1,..,S_{x}\}\times\{1,..,S_{y}\}$,
$W^{m,c}\in\mathbb{R}^{2^{K}}$,$T^c\in \mathbb{R}$ }}{\small \par}

\textbf{\small{{Output}}}{\small{{: A classifier decision in $\{1,..C\}$}}}{\small \par}

{\small{{Initialization: For $c=1,..,C ~~~~ Score[c]=-T^c $ }}}{\small \par}

{\small{{ $~~~~~~~~~~~~~~~$ Prepare extended image $I^{e}\in \mathbb{R}^{S_x \times S_y \times D_e}$ }} }{\small \par}

{\small{{For all tables $m=1,..,M$}}}{\small \par}

{\small{$~~~~~~$ For all pixels $p\in A^{m}$}}{\small \par}

{\small{{$~~~~~~~~~~~~$ Compute $F=B(I^e_{N(p)}; \Theta^m) \in {\{0,1\}}^K$ }}}{\small \par}


{\small{{$~~~~~~~~~~~~$ For $c=1,..,C ~~~ Score[c]=Score[c]+W^{m,c}[F]$ }}}{\small \par}

{\small{{Return $\arg\max_{c}Score[c]$ } }}} 
\end{algorithm}

We present the classifier structure in Section~\ref{sub:Struct} and derive the learning algorithm in~\ref{sub:Train}. The details and variations of structure and training appear in~\ref{sub:StructDetails} and \ref{sub:TrainDetails} respectively.

\subsection{Notation and classifier structure}
\label{sub:Struct}

A convolutional table ensemble is a classifier $f:I \to {\{1,..,C\}}$ where $I\in \mathbb{R}^{ S_x\times S_y \times D}$ and $C$ is the number of classes. The image $I$ may undergoes a preparation stage where additional feature channels ('maps') may be added to it, so it is transformed to $I^e \in \mathbb{R}^{ S_x\times S_y \times D_e}$ with $D_e\ge D$. 
After preparation, the ensemble sums the votes of $M$ convolutional tables, each in turn sums the votes of a word calculator over all pixels in a aggregation area. We now explain this process bottom-up.

\textbf{Word calculator:} For an image location $p=(x,y) \in \{1,..,S_x\} \times \{1,..S_y\}$, we denote
its neighborhood by $N(p)$, and by $I_{N(p)}$ the patch centered at location $p$. 
A word calculator is a feature extractor applied to such a patch and returning a $K$-bit index, i.e. a function $W:\mathbb{R}^{l\times l \times D_e} \to \{0,1\}^K$, where $l$ is the neighborhood size. In a classifier we have $M$ calculators denoted by $B^m = B(\cdot ~; \Theta^m)$ where $m=1,..,M$ and $\Theta^m$ are the calculator parameters. The calculator computes its output by applying $K$ bit-functions to the patch, each producing a single bit. 
Several types of bit functions are discussed in Section~\ref{sub:StructDetails}.

\textbf{ Convolutional table:} Each word calculator is applied to all locations in an integration area, and each word extracted casts votes for the $C$ output classes. Convolutional table $m$ is hence a triplet $(B^m,A^m,W^m)$ where $A^m \subset \{1,..,S_x\} \times \{1,..S_y\}$ is the integration area of word calculator $B^m$, and $W^m \in \mathbb{M}_{C \times 2^K }$ is its weight matrix. Denote by $b_p^m =B(I^e_{N(p)} ; \Theta^m)$ the word calculated at location $p$. We gather a histogram $H^m=(H^m_0, .., H^m_{2^K-1})$ counting word occurrences; i.e.,

\begin{equation}
H^{m}_b=\sum_{p\in A^{m}}\delta(b_{p}^{m}-b)     \label{eq:Table}
\end{equation}
with $\delta$ the discrete delta function. The class support of the convolutional table is the $C$-element vector $W^m (H^m)^t$.  

\textbf{ Convolutional tables ensemble:} The ensemble classification is done by accumulating the class support of all convolutional tables into a linear classifier with a bias term. Let $H = [ H^1, .., H^M]$ , and $W=[ W^1,..W^M] \in \mathbb{R}^{C \times M2^K}$. The classifier's decision is given by
\begin{equation}
C^{*} =  \arg\max_c W H^t - T^t    \label{eq:classifier}
\end{equation}
where $T=(T^1,..T^C)$ is a vector of class biases. Algorithm box~\ref{alg:CTE_test} shows the classifier's test time flow. Note that the histograms $H^m$ are not accumulated in practice, and instead each word computed directly votes for all classes. 

\begin{algorithm}[t]
\caption{\textbf{\label{alg:training}} Convolutional Tables Ensemble: Training}

{

\textbf{\small{{Input}}}{\small{{: A labeled training set $S=\{I_{i},y_{i}\}_{i=1}^{N}$}}}{\small \par}


{\small{{$~~~~~~~~~~~$ Parameters $M,K,C, {\{A^m\}}_{m=1}^M$ , convex loss $L(S ; W,T)$ }}}{\small \par}


\textbf{\small{{Output}}}{\small{{: A classifier $(\Theta^{m},A^{m},W^{m,c},T^c)_{m=1,c=1}^{M,C}$      }}}{\small \par}

{\small{{Initialization: $g_i^c=1/|\{I_{i}|y_{i}^c=1\}|~~$ if $y_{i}^c=1$,}}}{\small \par}

{\small{{$~~~~~~~~~~~~~~~~~~~~~~$ $g_i^c=-1/|\{I_{i}|y_{i}^c=-1\}|$
if $y_{i}^c=-1$}}}{\small \par}

{\small{{For $m=1,..,M$}}}{\small \par}

{\small{{$~~~~~~$ Table addition: choose  $\Theta^{m}$ to optimize: }}}{\small \par}

{\small{{$~~~~~~$    $\Theta^{m} = \arg\max_{\Theta}  \sum_c \sum_{b\in \{0,1\}^K} | \sum_{\{i,p: b_{i,p}(\Theta)=b\}} g_i^c |$  }}}{\small \par}

{\small{{$~~~~~~$ Update representation: $~~~ \forall i=1,..N, b\in{\{0,1\}}^K $  }}}{\small \par}

{\small{{$~~~~~~~~~~~~$  $H^m_b[I_i]= \sum_{p}\delta(b^m_{p,i}=b)$, $H=[ H, H^m ]$     }}}{\small \par}

{\small{$~~~~~~$   Global optimization:  train $W,T$ by solving  }}{\small \par}

{\small{$~~~~~~~~~~~~~$  $ \arg\min_{W,T} L( {\{H_i,y_i\}}_{i=1}^N ; W,T)$  }}{\small \par}

{\small{$~~~~~~$   If $m<M$ get loss gradients:  $ g_i^c = \frac{dL}{dS_i^c}  $    }}}{\small \par}

\end{algorithm}

\subsection{Training}
\label{sub:Train}

In~\cite{DFE,Ren_2014_CVPR} instances of convolutional tables ensemble were discriminatively optimized for specific tasks and losses (hand pose recognition using SVM in~\cite{DFE}, and face alignment using $l^2$ regression in~\cite{Ren_2014_CVPR}). The main idea behind these methods is to iterate between solving a convex problem for a fixed representation, and augmenting the representation based on gradient signals from the obtained solution. Here we adapt these ideas to linear $M$-classification with an  arbitrary $l_2$-regularized convex loss function, using techniques from~\cite{Mason2000,BarHillel2005}. Assume a labeled training sample with fixed representation ${\{(H_i,y_i)\}}_{i=1}^{N}$ where $H_i \in \mathbb{R}^{m2^K}$, $y_i \in \{1,..,C\}$, and denote the $c$-th row of the weight matrix $W$ by $W_c$. We want to learn a linear classifier of the form $C^{*}=\arg\max_c s^c$ with $s^c=W_c H^t-T^c$ by minimizing a sample loss function of the form 
\begin{equation}
L({\{H_i,y_i\}}_{i=1}^N) = \frac{1}{2}||W||^2 + \sum_{i=1}^N l({\{s_i^c\}}_{c=1}^C, y_i)    \label{eq:Loss}
\end{equation}

with $l({\{s^c\}}_{c=1}^C, y)$ a convex function of $s^c$. 
$L$ is strictly convex with a single global minimum, hence solvable using known techniques. Once the problem has been solved for the fixed representation $H_i$,
 we want to extend the representation by incorporating a new table, effectively adding $2^K$ new features. In order to choose the new features wisely, we consider how the loss changes if a new feature $f^{+}$ is added to the representation with small class weights, regarded as a small perturbation of the existing model.

Denote by $f^{+}_i$ the value of a new feature candidate for example $i$.After incorporating the new feature, example $i$'s representation changes from $H_i$ to $H_i^+ = [ H_i, f^{+}_i]$ and weights vectors $W_c$ are augmented to $[W_c, w^+_c]$ with $w^+_c\in \mathbb{R}$. Class scores $s_i^c$ are updated to $s_i^{c,+} = W^+_c (H_i^{+})^t -t^c = s_i^c + w^+_c f^{+}$. Finally, the loss is changed to $L^+ = L({\{H_i^+,y_i\}}_{i=1}^N) = \frac{1}{2}||W||^2 + \frac{1}{2}\sum_{c=1}^C {w^+_c}^2 + \sum_{i=1}^N l({\{s_i^{c,+}\}}_{c=1}^C, y_i)$. 
Denote $W^{+}=[w^+_1,..,w^+_C]$ the new weights vector. We assume that the new feature is added with small weights; i.e., $w^+_c\le \epsilon$ for all $c$. $L^+$ can be Taylor approximated around $W^{+}=0$, with the gradient $\frac{dL^+}{dW^{+}} |_{W^{+}=0}$:

\begin{equation}
\begin{aligned}
	\frac{dL^+}{dW^{+}} \bigg|_{W^{+}=0}  & = w^{+} + \sum_{i=1}^N \frac{dl({\{s_i^{c,+}\}},y_i)}{ds_i^{c,+}} f^{+}_i 	\bigg|_{W^{+}=0}   \\
	 &	=			\sum_{i=1}^N \frac{dl({\{s_i^{c}\}},y_i)}{ds_i^{c}} f^{+}_i  				\label{eq:LossGrad}
\end{aligned}
\end{equation}

Using the gradient in a Taylor approximation of $L^+$ gives
\begin{equation}
\begin{aligned}
	L^+  &= L + W^+ 	(\frac{dL^+}{dW^{+}})^t + O(||W^+||^2) \\			\label{eq:LossApprox}
	    &= L + \sum_{c=1}^C w^+_c \sum_{i=1}^N  \frac{dl({\{s_i^{c}\}},y_i)}{ds_i^{c}} f^{+}_i  + O(||W^+||^2)
\end{aligned}
\end{equation}

Denote $g_i^c = \frac{dl({\{s_i^{c}\}},y_i)}{ds_i^{c}}$. For loss minimization we want to minimize $\sum_{c=1}^C w^+_c \sum_{i=1}^N  g_i^c f^{+}_i$ over $W^+$ and $f^+$. For fixed $f^+$ minimizing over $W^+$ is simple. Denoting $R^c(f^+)=\sum_{i=1}^n g_i^c f_i^+$, we have to minimize $\sum_{c=1}^C w^+_c R^c(f^+)$ under the constraint $w^+_c\le \epsilon,~ \forall c$. We can minimize each term in the sum independently to get $w^{+,c}_{opt} = -\epsilon sign(R^c)$, and the value of the minimum is $-\epsilon \sum_{c=1}^C |R^c(f^+)|$. Hence, for a single feature addition, we need to maximize the score $\sum_{c=1}^C |R^c(f^+)|$.

To return to our scenario, we add $2^K$ features at once, generated by a new word calculator $B$. 
The derivation above can be done for each of them independently, so for the addition of the features ${\{H^+_b\}}_{b\in \{0,1\}^K}$ we get
\begin{equation}
\begin{aligned}
	L^+  & \approx L - \sum_{b \in {\{0,1\}}^K}  -\epsilon \sum_{c=1}^C |R^c(H^+_b) | \\ \label{eq:gradientAccum}
			 & =  L -  \epsilon \sum_{c=1}^C \sum_{b \in {\{0,1\}}^K}  \boldsymbol{|} \sum_{i=1}^N g_i^c \sum_{p \in A^+} \delta(b_{i,p}^+ = b)  \boldsymbol{|} \\
			 & = L - \epsilon \sum_{c=1}^C \sum_{b \in {\{0,1\}}^K}  \boldsymbol{|} \sum_{\{ i,p: b_{i,p}^+=b\}} g_i^c  \boldsymbol{|} ~\DeltaOverEq ~L -\epsilon R(B)
\end{aligned}
\end{equation}

where we used Equation~\ref{eq:Table} for $H^+_b$ and denoted $b_{i,p}^+=B(I_{i,N(p)})$. The resulting training algorithm, summarized in algorithm box~\ref{alg:training}, iterates between global classifier optimization and greedy optimization of the next convolutional table by maximizing $R(B; \Theta)$.

\subsection{Structural variants}
\label{sub:StructDetails}

The word calculator concept described in Section~\ref{sub:Struct} is very general. Here we describe the bit functions and word calculator types we have explored. 

\textbf{Bit functions and input preparation:} Word calculators compute an index descriptor of a patch $P\in \mathbb{R}^{l\times l \times D_e}$ by applying $K$ bit functions, each producing a single bit. Each such function is composed of a simple comparison operation, with a few parameters stating its exact operation.  Specifically we use the following bit function forms:


\begin{itemize}
\item {\bf One pixel:} $F(P) = \sigma(P(x,y,d)-t)$
\item {\bf Two pixels:} $F(P) = \sigma(P(x_1,y_1,d) - P(x_2,y_2,d)-t)$ 
\item {\bf Get Bit $l$:} $F(P) = ( P(0,0,d) << l ) \& 1$
\item {\bf Integral channel bit:} $F(p) = \sigma(P(x_1,y_1,d) - P(x_1,y_2,d) - P(x_2,y_1,d) + P(x_2,y_2,d)-t)$
\end{itemize}


where $\sigma$ is the Heaviside step function. The first two bit function types can be applied to any input channel $d\in \{1,..,D_e\}$, while the latter two are meaningful only for specific channels. The channels we consider are as follows:

\begin{itemize}
\item {\bf Original image channels:} Gray scale and color channels, or depth and IR (multiplied by a depth-based mask) for depth images.  
\item {\bf Gradient-based channels:} Two kinds of gradient maps are computed from the original channels following~\cite{Dollar2009}. A normalized gradient channel includes the norm of the gradient for each pixel location. In oriented gradient channels the gradient energy of a pixel is softly quantized into $N^O$ orientation maps.
\item {\bf Integral channels:} Integral images~\cite{Viola01} of channels from the previous two forms, again following~\cite{Dollar2009}. Applying integral channel bits to these channels allows fast calculation of channel area sums.
\item {\bf Spatial channels:} Two channels stating the horizontal and vertical location of a pixel in the image. These channels state the quantized location, using $N^H=\lfloor \log_2 S_x \rfloor$ and $ N^V=\lfloor \log_2 S_y \rfloor$ bits respectively.  
\end{itemize}


After preparation, the channels are optionally smoothed by a convolution with a triangle filter. Spatial channels enable the incorporation of patches' position in the word computed. They are used with a 'Get Bit $l$' bit function type, with $l$ referring to the higher bits. This effectively puts a spatial grid over the image, thus turning the global summation pooling into local summation using a pyramid-like structure~\cite{Lazebnik06}. For example using two bit functions, checking for the $N^H$-th horizontal bit and the $N^V$-th vertical bit, effectively puts a $2\times2$ grid over the image where words are summed independently and get different weights for each quarter. Similarly using $4$ spatial bits one gets a $4\times4$ pyramid, etc. We found that enforcing a different number of spatial bits in each convolutional table improves feature diversity and consequently the accuracy.

\textbf{Word calculator structure:} The main design decision in this respect is the choice between ferns and trees.
Ferns include only $K$ bit functions, so the number of parameters is relatively small and over-fitting during local optimization is less likely. Trees are a much larger hypothesis family
with up to $2^K-1$ bit functions in a full tree.
Thus they are likely to 
enable higher accuracy, but also be more prone to overfit. We explored this trade-off using a 'long tree' structure enabling a gradual interplay between the fern and full tree extremes. 

In a long tree the $K$ bits to compute are divided into $N^S$ stages, with $K_s$ bits computed at stage $s = 1,..,N^S$, so $\sum_{s=1}^{N^S} K_s =K$. A tree 
of depth $N^S$ is built, where a node in stage $s$ contains $K_s$ bit functions computing a $K_s$-bits word. 
A node in stage $s=1,..,N^S-1$ has $q_s$ children, and it has a child-directing table of size $2^{K_s}$, with entries containing child indices in $1,..,q_s$.
Computation starts at stage $1$ at a root node, and after computation of the $K_s$ bits in a node the produced word is used as an index to the child-directing table, whose output is the index of the child node to descend to. The tree structure is determined by the vectors $(K_1,..,K_{N^S})$ and $(q_1,..,q_{N^S-1})$ of stage size and stage split factors respectively. 

When speed is considered, the most important point is that ferns can be efficiently implemented using vector operations (like SSE), constructing the word in several locations at the same time. The efficiency arises because computing the same bit function for several contiguous patches involves access to contiguous pixels, which can be done without expensive gather operations. Conversely, for trees different bit functions are applied at contiguous patches 
so the accessed pixels are not contiguous in memory. As will be seen in Section~\ref{sec:EmpiricalResults}, trees can be more accurate, but ferns provide considerably better accuracy-speed trade-off.

\begin{figure*}
\begin{tabular}{| l |c |c| c| c| c |c| c| c| c | }
\hline
Dataset    &    DFE  & CTE base & \textbackslash Opt TH     & \textbackslash Ftr Norm      & \textbackslash WC opt & \textbackslash Chnls & \textbackslash Smooth & \textbackslash Spatial & \textbackslash Sp. Enforce \\
\hline
MNIST      &  0.77   & 0.45     & 0.48       & 0.58          & 0.48          & 0.7      & 0.66   &  0.51   & 0.48                \\
CIFAR-10   &  31.3   & 20.3     & 21.3       & 21.9          & 21.8          & 22.0     & 22.2   &  21.5   & 21.0               \\  
SVHN		   &  11.9   & 6.5		  & 7.1			   & 7.1           & 10.5				   & 11.6		  & 7.2	   & 13.2	  & 7.6				          \\ 
3-HANDPOSE &   3.2   & 2.3      & 2.1		  	 & 2.5           & 3.5					 & 4.4		  & 2.7 	 & 2.2	  & 2.2  	        \\ 
\hline
\end{tabular}
 \caption{\footnotesize \textbf{Comparison and ablation:} Columns one and two present errors of a DFE~\cite{DFE} and a baseline CTE. The next columns show results obtained when the CTE baseline is {\it \bf ablated} in a single aspect. \textbf{ Opt TH:} bit function thresholds are randomized instead of optimally chosen. \textbf{ Ftr Norm:} histogram features are not multiplicatively normalized before training. \textbf{ WC opt} bit functions in the word calculator are randomly chosen, and not optimized. \textbf{ Chnls:} appearance channels beyond original ones are not used. \textbf{ Smooth:} input channels are not smoothed during preprocessing. \textbf{ Spatial:} Spatial bits are not used. \textbf{ Sp. Enforce:} spatial bits are not enforced (When enforced, a random number in $\{1,..,5\}$ of spatial bits is enforced for each table). 
}
\label{table:Ablations} 
\end{figure*}

\subsection{Training variants}
\label{sub:TrainDetails}

As stated in algorithm~\ref{alg:training}, training iterates between gradient based word calculator optimization and global optimization of table weights. We now describe the methods we explored for these two components. 

\textbf{Word calculator optimization:} We consider several mechanisms for the optimization of $R(B;\Theta)$, including forward bit function selection, optimal threshold finding, and iterative bit function replacement/refinement.

In forward selection, we optimize $R(B)$ by adding one bit after the other. For fern growing there are $K$ such stages. At stage $l=1,..K$, ${\{F^j\}}_{j=1}^{N_c}$ candidate bit functions are generated, with their type and parameters drawn from a prior distribution. For each $j$, we augment the current word calculator $B$ to $B^+=[B,F^j]$ and choose the one with the highest score. However, we found that simple greedy computation of $R(B^+)$ at each stage is not the best way to optimize $R(B)$, and an auxiliary score which additively normalizes the newly-introduced features does a better job. Denote the patch features of a word calculator $B$ by $\delta^b(P)=\delta(B(P)=b)$, by $\delta^b_{i,p}$ the value of $\delta^b$ for pixel $p$ in image $i$ and by $R^c(f(P)) \DeltaOverEq \sum_{i,p} g^c_i f_{i,p}$ the score $R$ induced by a patch feature $f$. The addition of a new bit effectively replaces the feature $\delta^b$ for $b\in\{0,2^l-1\}$ with $2$ new features $\delta^{(b,0)}$ and $\delta^{(b,1)}$. If the gradients in cell $b$ are not balanced; i.e., $\sum_{i,p} g^c_i \delta^b_{i,p}= C_0 \ne 0$, as is often the case, a feature $\delta^{(b,0)}$ may get a good $R^c(\delta^{(b,0)})$ score of $C_0$ even if the new bit function is constant, or otherwise uninformative. To handle this, we score a normalized version of the new features, with an average value of 0, which more effectively measures the added information in the new features. The following lemma shows that this is a valid, as well as computationally effective strategy:

\begin{lemma}\label{lemma:L1Delta} Let $\bar{\delta}^{(b,u)} = \delta^{(b,u)} - \frac{\#\delta^{(b,u)}}{\#\delta^b}\delta^{b}$ for $u=0,1$ and $\#\delta^a = \sum_{i,p} \delta^a_{i,p}$. The following properties hold
\begin{enumerate}
\item \label{lem:equivalence} Using $\bar{\delta}^{(b,1)}$, $\delta^b$ in a classifier is equivalent to using $\delta^{(b,0)}$,$\delta^{(b,1)}$; i.e, for any weight choice $w_0,w_1$ there are $w_b,w_\Delta$ such that $w_0 \delta^{(b,0)} + w_1 \delta^{(b,1)} = w_b\delta^b + w_\Delta \bar{\delta}^{(b,1)}$
\item \label{lem:half}  $R^c(\bar{\delta}^{(b,0)}) = R^c(\bar{\delta}^{(b,1)})$ 

\item \label{lem:dual}  $R^c(\bar{\delta}^{(b,0)})= \sum_{i,p}(g^c_i - E[g^c_i|b])\delta^{(b,0)}$ with $E[g^c_i|b]  \DeltaOverEq \frac{\sum_{i,p}g^c_i\delta^b}{\#\delta^b}$
\end{enumerate}
\end{lemma}

The proofs are rather simple and appear in appendix~\ref{App:LemmaProof}. 
Property~\ref{lem:equivalence} shows that we may score $\delta^b,\bar{\delta}^{(b,1)}$ features instead of $\delta^{(b,u)}$ features. Since only $\bar{\delta}^{(b,1)}$ is affected by the new candidate bit, we can score only those terms when selecting among candidates. 
Property~\ref{lem:dual} shows that we can normalize the gradient instead of the feature candidates, which is cheaper (as there are $N_c$ candidates but only a single gradient vector). In summary, we optimize the next bit selection by maximizing
\begin{equation}
\label{eq:L1Delta}
R_\Delta ([B,F^j]) \DeltaOverEq \sum_{c=1}^C \sum_{ b\in \{0,1\}^{l}}  \boldsymbol{|} \sum_{\{i,p: b_{i,p}=(b,1)\}} (g_i^c - E[g^c_i|b]) \boldsymbol{|}
\end{equation}

over the choice of $F^j$. The calculation requires a single histogram aggregation sweep over all patches $(i,p)$.
 
Most of the bit functions obtain their bit by comparing an underlying patch measurement to a threshold $t$. For such functions, the optimal threshold parameter can be found with a small additional computational cost. This is done by sorting the underlying values of $F^j$ and computing the sum over $i,p$ in Equation~\ref{eq:L1Delta} by running with the sorted order. This way, a running statistic of the $R_{\Delta}$ score can be maintained, computing the score for all possible thresholds and keeping the best.

For a long tree a similar algorithm is employed, with ferns internal to nodes optimized as full ferns, but tree splits requiring special treatment. Assume we are splitting a node in stage $s$, so the current word calculator has already computes a $L_s=\sum_{i=1}^s K_i$-bit word, among which $K_s$ were computed in the current node. We now choose the first bit functions of all the children, as well as the redirection table, to optimize $R$. Since different prefixes of the current calculator $B$ are augmented by different bit functions we need to decompose the $R$ score. Denote by $a$ the index set $\{L_s-K_s+1,..,L_s\}$ of bits computed by the current node, and by b(a) the limitation of a binary word $b$ to indices $a$. 
For a $K_s$-bit word $z\in{\{0,1\}}^{K_s}$, we define the component of $R$ contributed by words with $b(a)=z$ by

\begin{equation}
R_{b(a)=z}(B) \DeltaOverEq \sum_{c=1}^C \sum_{ \substack{b\in \{0,1\}^{L_s} \\ b(a)=z} } \boldsymbol{|} \sum_{\{i,p: b_{i,p}=b\}} g_i^c \boldsymbol{|}
\end{equation}


For the tree split we draw a large set $\mathbb{F}$ of candidate bits, and choose the first bits of the $q_s$ children by optimizing

\begin{equation}
\max_{ \substack{G \subset \mathbb{F} \\ |G|=q_s}}  \sum_{z \in {\{0,1\}}^{K_s}} \max_{F_{z}\in G} R_{b(a)=z}([B,F_{z}])
\end{equation}

with $G$ the set of chosen bits for the children and entry $z$ in the redirection table set to the index of the child containing $F_z$. For this optimization we compute the score matrix $S \in \mathbb{M}_{2^{K_s} \times |\mathbb{F}|}$ with $S(i,j)=R_{b(a)=i}([B,F_j])$. Given a choice of $G$, amounting to a choice of column subset in $S$, the optimization over $F_z$ is trivial and the score is easy to compute. We optimize over $G$ by exhaustively trying all choices of $G$ for $|G|=2$, and greedily adding columns to $G$ until it contains $q_s$ members.  

In addition to forward bit selection, we implemented iterative bit replacement and refinement stages. The rationale for this is the observation that while the last bit functions in a fern are chosen to complement the previous ones, the bits chosen at the beginning are not optimized to be complementary and may be suboptimal in a long word calculator. The bit replacement algorithm operates after forward bit selection. It runs over all the bit functions several times and attempts to replace each function with several randomly drawn candidates. A replacement step is accepted if it improves the $R_{\Delta}(B)$ score. In a similar manner, a bit refinement algorithm attempts to replace a bit function by small perturbations of its parameters, thus effectively implementing a local search. For trees, bit replacement/refinement is done only for bits inside a node, and once a split is made the node parameters are fixed.

\begin{figure*}
\begin{tabular}{cc}
\begin{tabular}{| l | c | c | c |}
\hline
Dataset and Error & CTE ($\mu S$) & CNN ($\mu S$) & speedup   \\
\hline
MNIST 0.01		  		 &  4.8  			 &  63.9  		 & 13.1$\times$      \\
CIFAR-10 0.25   		 & 168.8 			 &  882  		   & 5.2$\times$     \\
SVHN  0.15 		  		 & 18.4  			 &  88  		   & 4.7$\times$     \\
3-HANDPOSE 0.035		 & 6.3				 &  1250       & 199.3$\times$		 \\
\hline
\end{tabular}  &
\begin{tabular}{| l | c | c | c |}
\hline
Dataset &  Tree form   								 & Error(\%)  & speed ($\mu S$) \\ 
\hline
MNIST    &  Fern				 					     & 0.45    		& 106           \\
				 &  Depth 3, 4-way splits		   & 0.43  	  	& 398           \\
				 &  Depth 6, 2-way splits 	   & 0.39  	  	& 498           \\
\hline
CIFAR    &  Fern  										 & 21.0  			& 800         \\
	~~-10	 &  Depth 3, 4-way splits			 & 19.7   		& 3111        \\
				 &	Depth 4, 3-way splits      & 19.3       &	3544			  \\
\hline
\end{tabular}
\end{tabular}
 \caption{\footnotesize \textbf{Left:CTE-CNN speed differences:} Some examples of required accuracy points where a CTE can meet the accuracy while being considerably faster than a CNN. 
\textbf{Right:Ferns/Trees trade-off:} Accuracy and speed for several fern/tree configurations on MNIST and CIFAR-10. The fern classifiers are the baseline classifiers whose results are reported in Figure~\ref{table:Ablations}. For trees the bits are split evenly between the $N^S$ stages (for MNIST, which uses $11$-bit tables, the last stage gets one less bit), and all the split parameters are equal.
}
\label{fig:MoreTables} 
\end{figure*}

\textbf{Global optimization:}
We considered two global loss functions in our classification experiments: an SVM-based loss, and a softmaxloss as typically used in neural networks optimization. In the SVM loss, we take the sum of $C$ SVM programs, each minimizing a one-versus-all error. Let $y_{i,c}=2*\delta(y_{i},c)-1$ be binary class labels. The loss is

\begin{equation}
L_{SVM} = \frac{1}{2}||W||^2 + \Lambda \sum_{c=1}^C \sum_{i=1}^N max(1-y_{i,c}s^c_i, 0)
\end{equation}

The loss aims for class separation in $C$ independent classifiers. Its advantage lies in the availability of fast and scalable methods for solving large and sparse SVM programs~\cite{Shalev07,Hsieh2008}. The loss gradients are $g^c_i = -y_{i,c}$ if example $i$ is a support vector, and $0$ otherwise. In~\cite{BarHillel10} a first order approximation for $\min_W L_{SVM}$ is derived for new feature addition, in which the example gradients are $-\alpha_i y_{i,c}$ with $\alpha_i$ the dual SVM variables at the optimum. The two expressions are similar and we did not find noticeable difference between them empirically. 
The softmax loss is 
\begin{equation}
L_{LR} = \frac{1}{2}||W||^2 - \Lambda \sum_{i=1}^N \log \frac{exp(s_i^{y_i})}{\sum_c exp(s^c_i)}
\end{equation}
This loss provides a direct minimization of the $M$-class error. 
The gradients are $g^c_i=exp(s_i^{y_i})/ \sum_c exp(s^c_i) - \delta(y_{i},c)$.
Conveniently, it can be extended to a distillation loss~\cite{Hinton2015}, which enables guidance of the classifier using an internal representation of a well-trained CNN classifier. 

Features in a word histogram have significant variance, 
as some words appear in large quantities in a single image. Without normalization such words may be arbitrarily preferred due to their lower regularization cost- they can be used with lower weights. Denote the column of a feature across all examples by $Col^m_b= (H^m_{b,1},..H^m_{b,N})$. We found that normalizing each features column by the expected count of active examples $L_1(Col^m_b)/ L_0(Col^m_b)$  improved accuracy and convergence speed in many cases.


\section{Empirical results}
\label{sec:EmpiricalResults}

We discuss our experimental setup in~\ref{sub:Setup}. In Section~\ref{sub:Ablations} we compare to related art and evaluate the contribution of algorithmic components to the performance. Results of speed-accuracy trade-offs are presented in~\ref{sub:AS}.

\subsection{Implementation and data details}
\label{sub:Setup}

The experiments were conducted on $4$ publicly available datasets: MNIST, CIFAR-10, SVHN and 3-HANDPOSE. The first three are standard recognition benchmarks in gray-scale (MNIST) or RGB (CIFAR-10,SVHN), with $10$ classes each. 3-HANDPOSE are a $4$-class dataset, with $3$ hand poses and a fourth class of 'other', and its images contain depth and IR channels. The image sizes are between $28 \times 28$ (MNIST) and $36 \times 36 $ (3-HANDPOSE). The training set size ranges from $50000$ (CIFAR-10) to $604000$ (SVHN). 

CTE training code was written in Matlab, with some routines using code from the packages~\cite{PMT2006,Fan2008,vedaldi15matconvnet}. The test time classier was implemented and optimized in C.
For ferns we implemented algorithm~\ref{alg:CTE_test} with SSE operations. Words are computed for $8$ neighboring pixels together, and voting is done for $8$ classes at once.
For trees
we implemented a program generating efficient code of the bit computation loop for a specific tree, so the tree parameters are part of the code. This obtained an acceleration factor of ~$2X$ over standard C code.
We also thread-parallelized the algorithm over the convolutional tables, with good a speed-up of $3.6\times$ obtained from $4$ cores. However, we report and compare single thread performance to keep the methodology as simple as possible.

CNN models were trained using MatConvNet~\cite{vedaldi15matconvnet}. The implementation is efficient, reported to be comparable to Caffe~\cite{Caffe2014} in~\cite{vedaldi15matconvnet}, with the convolutional and global layers reduced to matrix multiplication done using an SSE-optimized BLAS package. When measuring execution time, we measured net run time of the convolutional, pooling and global layers alone, without Matlab overhead.  
Time measurements were made on a Lenovo Thinkpad W530 quad core laptop, with i7-3720QM core running at 2.6Ghz. 

\begin{figure*}
\begin{tabular}{cccc}
\includegraphics[scale=0.09]{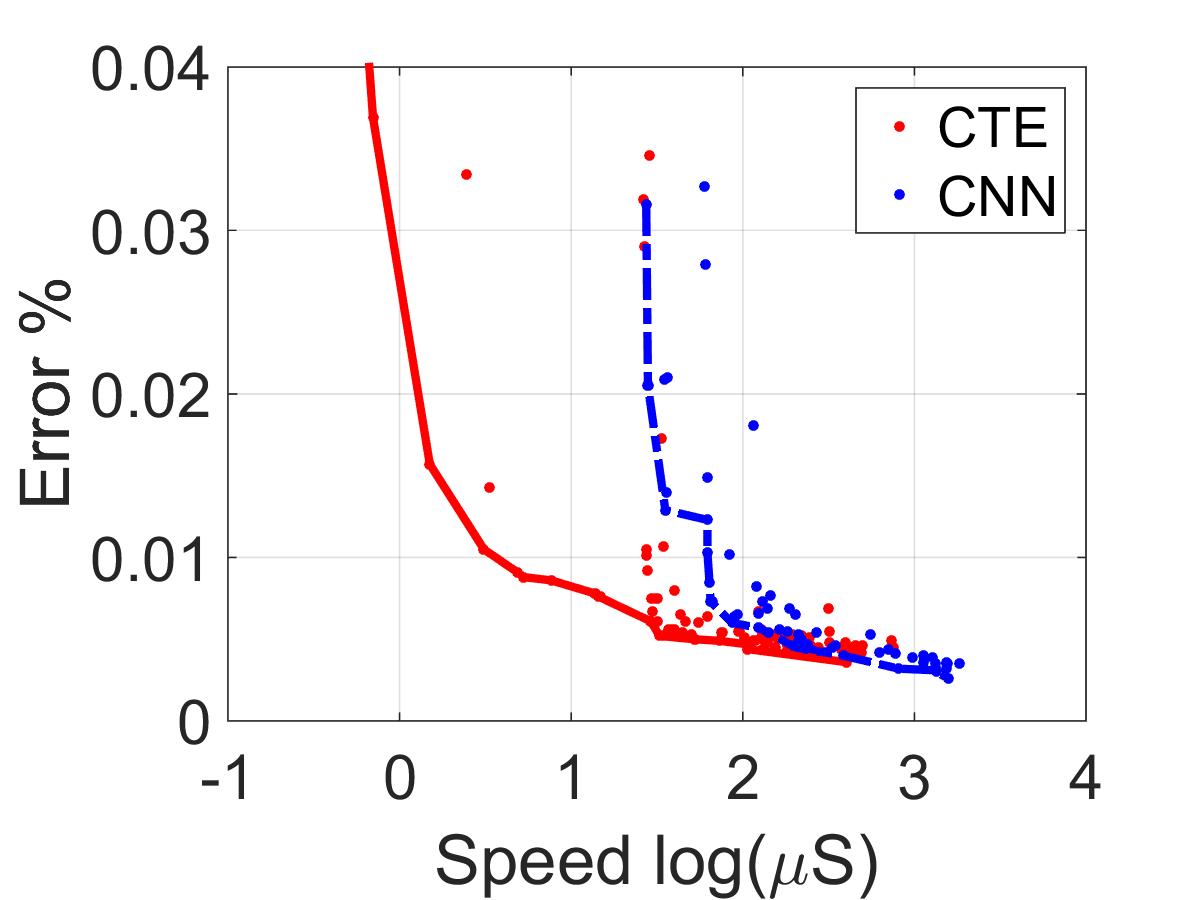}  &
\includegraphics[scale=0.09]{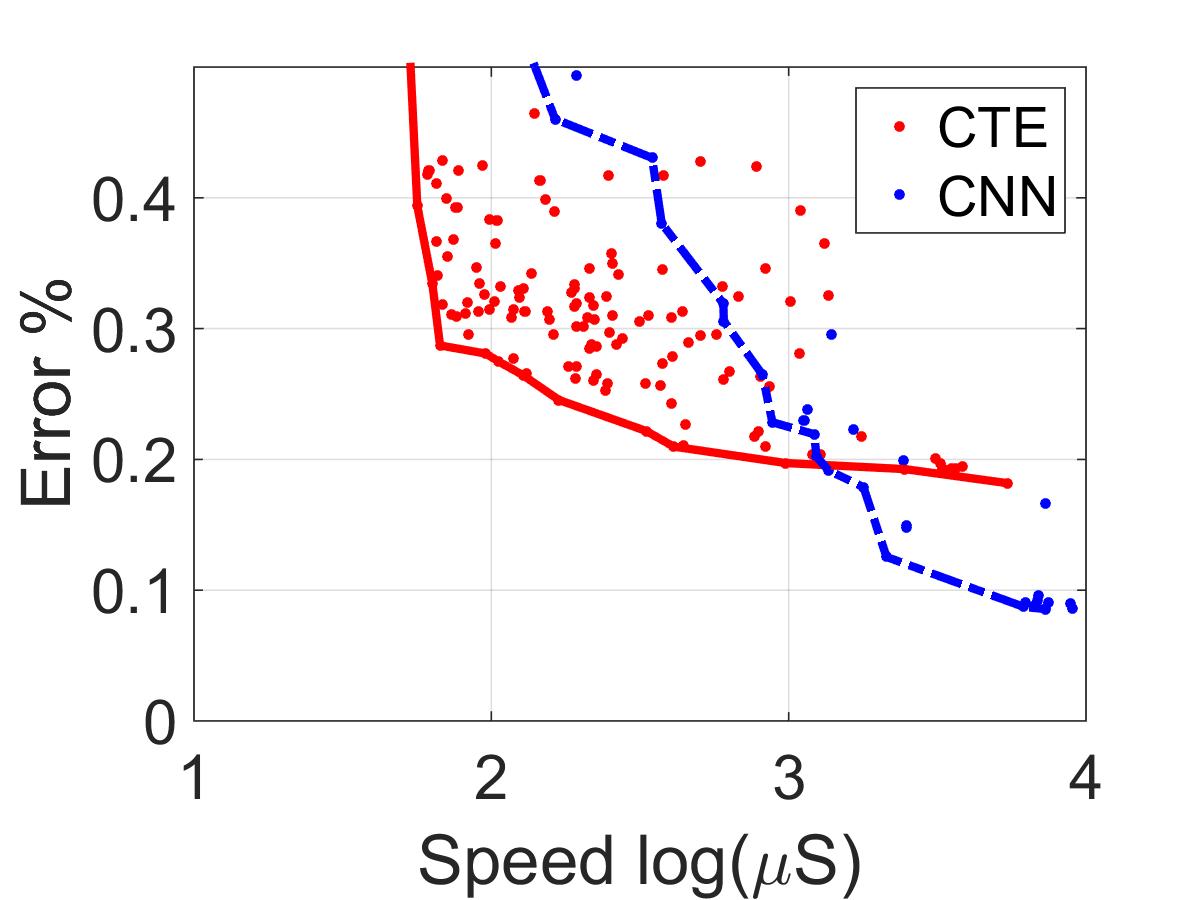}     & 
\includegraphics[scale=0.09]{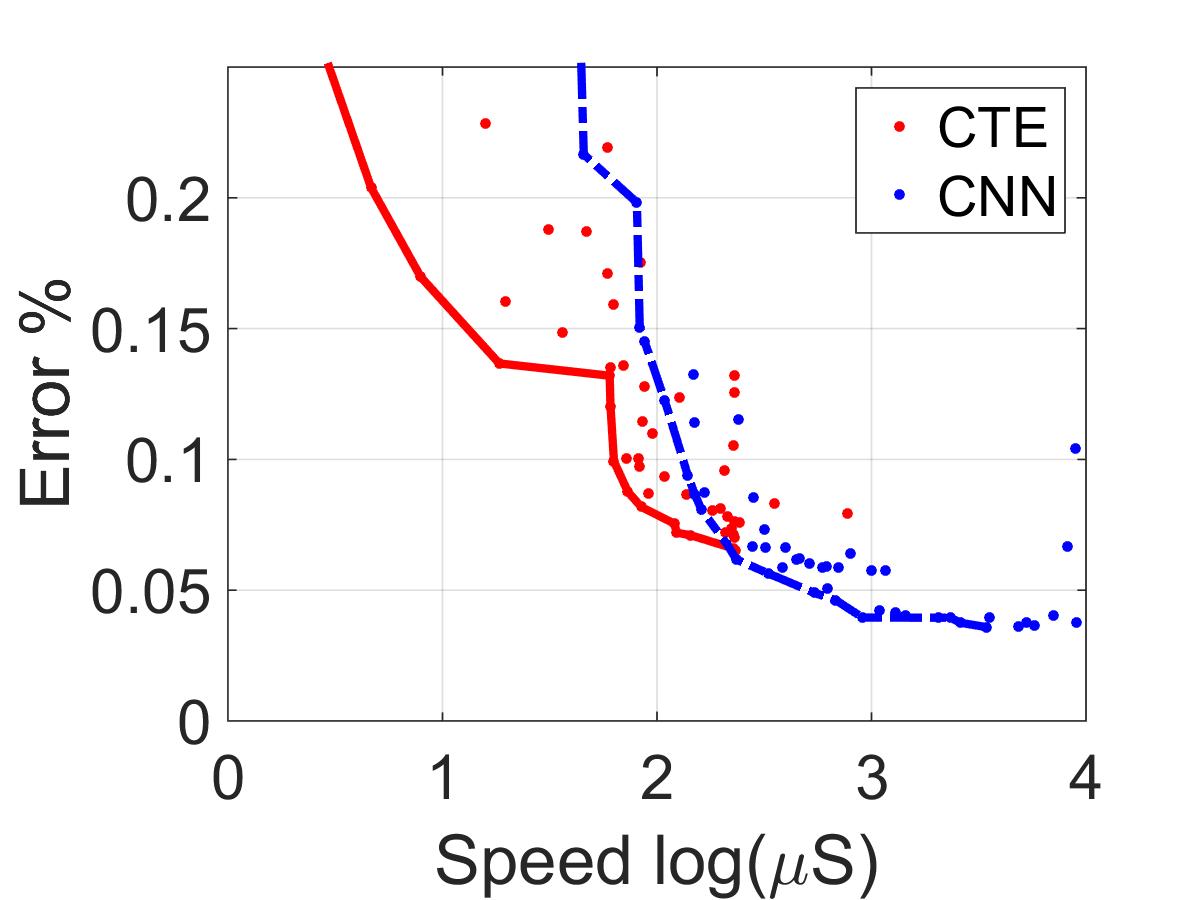}  &
\includegraphics[scale=0.09]{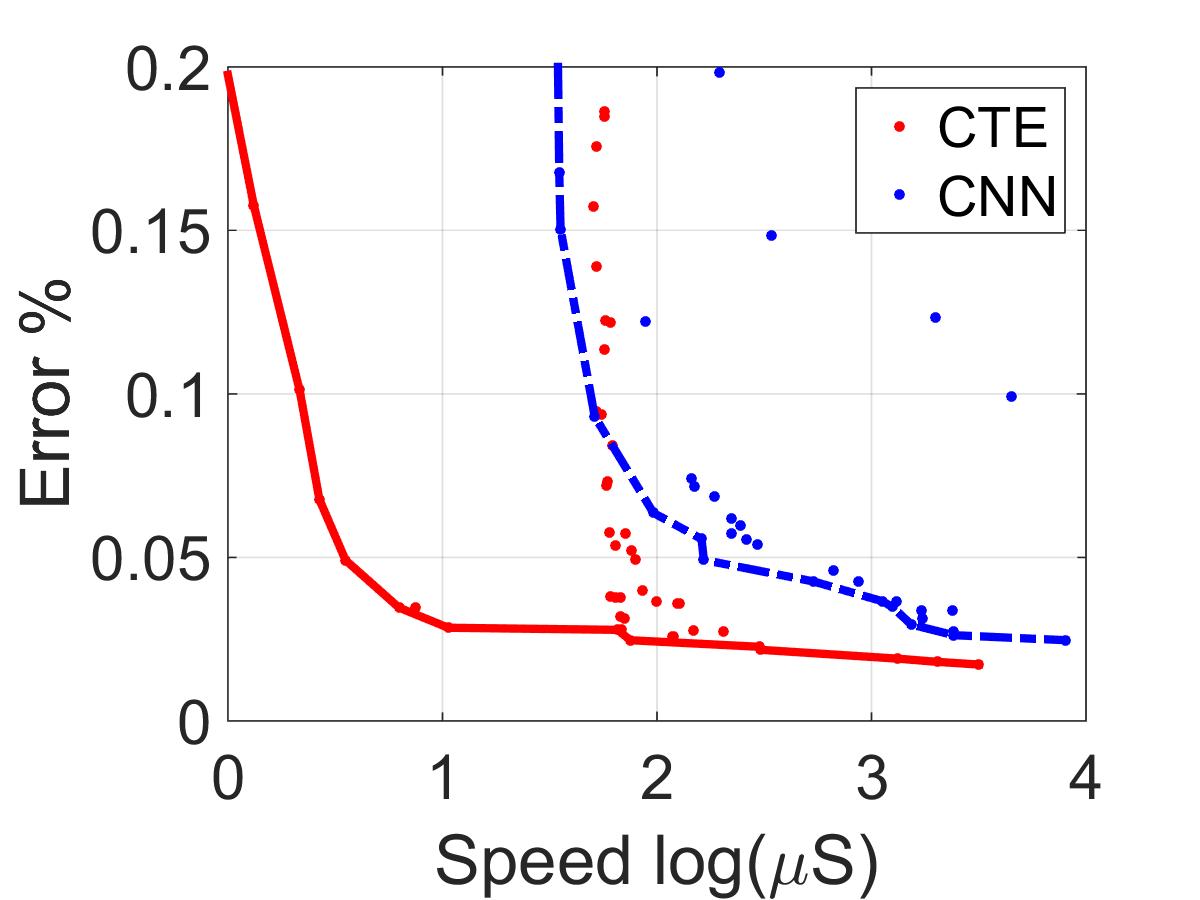}     \\ 
MNIST & CIFAR-10 & SVHN & 3-HANDPOSE 
\end{tabular}
 \caption{\footnotesize \textbf{Speed-accuracy trade-offs curves.} 
X axis states classifier run time in microseconds, in log scale base $10$. Each point states the (speed,acuarcy) result of a single trained classifier. the Lines are the lower envelope of classifiers from the CTE and CNN family. The number of classifiers trained over the data is 172 (MNIST), 283 (CIFAR-10), 89 (SVHN) and 111 (3-HANDPOSE).
}
\label{fig:TradeOffs} 
\end{figure*}

\subsection{Comparison and variation}
\label{sub:Ablations} 

\textbf{Comparison with DFE:} The Discriminative Ferns Ensemble (DFE) was suggested in~\cite{DFE} for classification of 3-HANDPOSE, and can be seen a baseline for CTE, which enhances it in many aspects. The first two columns in Table~\ref{table:Ablations} present errors of DFE and CTE on the $4$ datasets, using $50$ ferns for MNIST, SVHN, 3-HANDPOSE and $100$ for CIFAR-10. MNIST was trained with softmax distillation loss (see below for details), and the others with SVM loss. The aggregation area ${\{A^m\}}_{m=1}^M$ were chosen to be identical for all tables in a classifier, forming a centered square occupying most of the image.
To enable the comparison, M-class error rates are extracted from DFE (in~\cite{DFE} such errors are not reported, and $3$-class average true positive rates are reported instead). It can be seen that CTE base provides significant improvements of $24-45\%$ error reduction over DFE, with $28\%$ obtained for 3-HANDPOSE, where DFE was originally applied. Note that the CTE base is not the best choice for 3-HANDPOSE. With additional parameter tuning result of $2\%$ can be obtained with $50$ ferns, which is an improvement of $38\%$ over DFE.

\textbf{Ablation experiments:} The accuracy obtained by a CTE is influenced by many small incremental improvements related to structural and algorithmic variations. Columns 3-9 in Table~\ref{table:Ablations} show the contribution of some ingredients by removing them from the baseline CTE. For MNIST, where the effects are small due to the low error, results were averaged over $5$ experiments varying in their random seed, a with seed-induced std of $0.1\%$. It can be seen that these ingredients consistently contribute to accuracy for non-depth data. 

\textbf{Trees/Ferns trade-off:} The trade-off between ferns and trees for MNIST and CIFAR-10 is presented in Figure~\ref{fig:MoreTables}(Right). For MNIST, the results were averaged over $5$ experiments, with a seed induced std of $0.028\%$. It can be seen that trees provide better accuracy. However, the speed cost of using trees is significant, due to the inability to efficiently vectorize their implementation.   

\textbf{Distillation experiments:} We experimented with knowledge distillation from a CNN to a CTE using the method suggested in~\cite{Hinton2015}. In such experiments, soft labels are taken from our best CNN model, and a CTE is trained to optimize a convex combination of the standard softmax loss and the Kullback-Leibler distance from the CNN-induced probabilities.
We attempted this for MNIST and CIFAR-10 using our best CNN models, providing $0.31\%$ and $8.6\%$ error  respectively as distillation sources. For MNIST, this training methodology proved to be successful. Averaging over $5$ seeds, the accuracy of a $50$-fern CTE optimized for softmax was $0.66\%$(the std was $0.025\%$) without distillation, and $0.45\%(0.029\%)$ with distillation. For comparison, an SVM-optimized CTE with the same parameters obtained $0.61\%(0.04\%)$ error. For CIFAR-10 distillation did not consistently improve the results.


\subsection{Speed-Accuracy trade-off}
\label{sub:AS}

We are interested in the trade-off or Pareto curves~\cite{Climacao1997}, showing the best accuracy obtainable for a specific speed constraint and vice versa. Since the design space for variations of $CNNs$ and $CTE$ algorithms is huge, and the training time of the algorithms is considerable, we needed to sample it wisely to get a good curve approximation. Our sampling technique is based on two stages. In stage 1 we searched for the most accurate classifiers for CTE and CNN with loose speed constraints, so even slow classifiers 
were considered. We then used the few top accuracy variants of each architecture as baselines and accelerated them by systematically varying certain design parameters.  

Our CNN baseline architectures are variations of DeepCNiN(l,k)~\cite{Graham2014}, with $l=3-4$ convolutional layers and $k=60-100$, implying usage of $i\cdot k$ maps at the $i$-th layer. It was shown in~\cite{Graham2014} that higher $l,k$ values provide better accuracy, but such architectures are much slower than $1$ CPU millisecond and so they are outside our domain of interest. We experimented with dropout~\cite{Srivastava2014}, parametric RELU units~\cite{He2015}, affine image transformations following~\cite{Graham2014}, and HSV image transformations following~\cite{Snoek2015}. Acceleration of the baseline architectures used three main parameters. The first was reducing parameter $k$ controlling the network width. The second was reduction of the number of maps in the output of the NIN layers. This reduces the number of input maps for the next layer, and can dramatically save computation with relatively small loss of accuracy. The third was raising the convolution stride parameter from $1$ to $2$. For CTEs, our exploration space was sketched in Section~\ref{sec:Method}, and it includes both ferns and trees. The best performing configurations were then accelerated using a single parameter: the number of tables in the ensemble. 

Trade-off graphs for the $4$ datasets are shown in Figure~\ref{fig:TradeOffs}. Classification speed in microseconds is displayed along the $X$-axis in log scale with base $10$. For all datasets, there is a high speed regime where CTEs provide better accuracy than CNNs. Specifically CTEs are preferable for all datasets when less than $100$ microseconds are available for computation. 
Starting from $1000$ microseconds and up CNNs are usually better, with CTEs still providing comparable accuracy for MNIST and 3-HANDPOSE at the $1-10$ milliseconds regime.
Viewed conversely, for a wide range of error rates, if the error rate is obtainable by a CTE, it is obtainable with significant speedups over CNNs. Some examples of this phenomenon are given in Figure~\ref{fig:MoreTables}(Left). Note that while a working point of $0.25$ error for CIFAR-10 may seem high, the majority of the one-versus-one errors of such a classifier are lower than $0.05$, which may be good enough for many purposes.

\section{Conclusions and further work}
\label{sec:Conclusions}

We introduced improvements to the convolutional tables framework in terms of bit functions used, word calculator structure, calculator optimization and global optimization. We have shown that for highly computational constrained tasks CTE may provide accuracy higher than CNNs. A natural direction for future research is to replace the flat structure of CTEs with a layered approach, in order to try and enjoy the accuracy of CNNs with the speed of CTEs.
  
{\small
\bibliographystyle{ieee}
\bibliography{CTEbib}

\begin{thebibliography}{10}\itemsep=-1pt

\bibitem{Liu_2015_CVPR}
L.~Baoyuan, W.~Min, F.~Hassan, T.~Marshall, and P.~Marianna.
\newblock Sparse convolutional neural networks.
\newblock In {\em CVPR}, 2015.

\bibitem{BarHillel2005}
A.~Bar-Hillel, T.~Hertz, and D.~Weinshall.
\newblock Object class recognition by boosting a part based model.
\newblock In {\em CVPR}, 2005.

\bibitem{BarHillel10}
A.~Bar-Hillel, D.~Levi, E.~Krupka, and C.~Goldberg.
\newblock Part-based feature synthesis for human detection.
\newblock In {\em ECCV}, 2010.

\bibitem{Benenson2012}
R.~Benenson, M.~Mathias, R.~Timofte, and L.~J.~V. Gool.
\newblock Pedestrian detection at 100 frames per second.
\newblock In {\em CVPR}, 2012.

\bibitem{Bosch07}
A.~Bosch, A.~Zisserman, and X.~Mu\~noz.
\newblock Image classification using random forests and ferns.
\newblock In {\em ICCV}, pages 1--8, 2007.

\bibitem{RF_Cambridge}
A.~Criminisi, J.~Shotton, and E.~Konukoglu.
\newblock Decision forests for classification, regression, density estimation,
  manifold learning and semi-supervised learning.
\newblock Technical report, Microsoft Research, 2011.

\bibitem{DPM_LSH13}
T.~Dean, M.~Ruzon, M.~Segal, J.~Shlens, S.~Vijayanarasimhan, and J.~Yagnik.
\newblock Fast, accurate detection of 100,000 object classes on a single
  machine.
\newblock In {\em Proceedings of IEEE Conference on Computer Vision and Pattern
  Recognition}, Washington, DC, USA, 2013.

\bibitem{PMT2006}
P.~Doll\'ar.
\newblock {P}iotr's {C}omputer {V}ision {M}atlab {T}oolbox ({PMT}).
\newblock http://vision.ucsd.edu/~pdollar/toolbox/doc/index.html.

\bibitem{Dollar2009}
P.~Dollar, Z.~Tu, P.~Perona, and S.~Belongie.
\newblock Integral channel features.
\newblock In {\em BMVC}, 2009.

\bibitem{Climacao1997}
C.~J. (Ed).
\newblock {\em Multicriteria Analysis}.
\newblock Springer-Verlag, 1997.

\bibitem{Fan2008}
R.-E. Fan, K.-W. Chang, C.-J. Hsieh, X.-R. Wang, and C.-J. Lin.
\newblock Liblinear: A library for large linear classification.
\newblock {\em Journal of Machine Learning Research}, 9:1871--1874, 2008.

\bibitem{Geurts2006}
P.~Geurts, D.~Ernst, and L.~Wehenkel.
\newblock Extremely randomized trees.
\newblock {\em Mach. Learn.}, 63(1):3--42, Apr. 2006.

\bibitem{Graham2014}
B.~Graham.
\newblock Spatially-sparse convolutional neural networks.
\newblock {\em CoRR}, abs/1409.6070, 2014.

\bibitem{He2013}
K.~He, S.~Zhang, X.and~Ren, and J.~Sun.
\newblock Spatial pyramid pooling in deep convolutional networks for visual
  recognition.
\newblock {\em CoRR}, abs/1406.4729v2, 2014.

\bibitem{Hinton2015}
G.~Hinton, O.~Vinyals, and J.~Dean.
\newblock Distilling the knowledge in a neural network.
\newblock {\em CoRR}, abs/1503.02531, 2015.

\bibitem{Hsieh2008}
C.-J. Hsieh, K.-W. Chang, C.-J. Lin, S.~S. Keerthi, and S.~Sundararajan.
\newblock A dual coordinate descent method for large-scale linear svm.
\newblock In {\em ICML}, 2008.

\bibitem{Ba2014}
B.~Jimmy and C.~Rich.
\newblock Do deep nets really need to be deep?
\newblock In {\em NIPS}, 2014.

\bibitem{He2015}
H.~Kaiming, Z.~Xiangyu, R.~Shaoqing, and S.~Jian.
\newblock Delving deep into rectifiers: Surpassing human-level performance on
  imagenet classifcation.
\newblock {\em CoRR}, abs/1502.01852, 2015.

\bibitem{Krizhevsky2009}
A.~Krizhevsky and G.~Hinton.
\newblock Learning multiple layers of features from tiny images.
\newblock Technical report, Master’s thesis, Department of Computer Science,
  University of Toronto, 2009.

\bibitem{DFE}
E.~Krupka, A.~Vinnikov, B.~Klein, A.~B. Hillel, D.~Freedman, and S.~Stachniak.
\newblock Discriminative ferns ensemble for hand pose recognition.
\newblock In {\em CVPR}, 2014.

\bibitem{Lazebnik06}
S.~Lazebnik, C.~Schmid, and J.~Ponce.
\newblock Beyond bags of features: Spatial pyramid matching for recognizing
  natural scene categories.
\newblock In {\em CVPR}, pages 2169--2178, 2006.

\bibitem{MNist1998}
Y.~LeCun, L.~Bottou, Y.~Bengio, and P.~Haffner.
\newblock Gradient-based learning applied to document recognition.
\newblock In {\em Proceedings of the IEEE}, 1998.

\bibitem{Lepetit2006}
V.~Lepetit and P.~Fua.
\newblock Keypoint recognition using randomized trees.
\newblock {\em PAMI}, 28:1465--1479, 2008.

\bibitem{AFS2013}
D.~Levi, S.~Silberstein, and A.~Bar-Hillel.
\newblock Fast multiple-part based object detection using kd-ferns.
\newblock In {\em CVPR}, 2013.

\bibitem{Lin2013_NIN}
M.~Lin, Q.~Chen, and S.~Yan.
\newblock Network in network.
\newblock {\em CoRR}, abs/1312.4400, 2014.

\bibitem{Mason2000}
L.~Mason, J.~Baxter, P.~Bartlett, and M.~Frean.
\newblock Boosting algorithms as gradient descent.
\newblock In {\em NIPS}, pages 512--518, 2000.

\bibitem{Netzer2011}
Y.~Netzer, T.~Wang, A.~Coates, A.~Bissacco, B.~Wu, , and A.~Y. Ng.
\newblock Reading digits in natural images with unsupervised feature learning.
\newblock In {\em NIPS Workshop on Deep Learning and Unsupervised Feature
  Learning}, 2011.

\bibitem{Ren_2014_CVPR}
S.~Ren, X.~Cao, Y.~Wei, and J.~Sun.
\newblock Face alignment at 3000 fps via regressing local binary features.
\newblock In {\em CVPR}, June 2014.

\bibitem{Romero2015}
A.~Romero, N.~Ballas, S.~E. Kahou, A.~Chassang, C.~Gatta, and Y.~Bengio.
\newblock Fitnets: Hints for thin deep nets.
\newblock {\em CoRR}, abs/1412.6550, 2015.

\bibitem{Shalev07}
S.~Shalev-Shwartz, Y.~Singer, and N.~Srebro.
\newblock Pegasos: Primal estimated sub-gradient solver for svm.
\newblock In {\em ICML}, 2007.

\bibitem{ShottonSKFFBCM13}
J.~Shotton, T.~Sharp, A.~Kipman, A.~W. Fitzgibbon, M.~Finocchio, A.~Blake,
  M.~Cook, and R.~Moore.
\newblock Real-time human pose recognition in parts from single depth images.
\newblock {\em Commun. ACM}, 56(1):116--124, 2013.

\bibitem{Snoek2015}
J.~Snoek, O.~Rippel, K.~Swersky, R.~Kiros, N.~Satish, N.~Sundaram, M.~A.
  Patwary, Prabhat, and R.~P. Adams.
\newblock Scalable bayesian optimization using deep neural networks.
\newblock {\em CoRR}, abs/1502.05700, 2014.

\bibitem{Srivastava2014}
N.~Srivastava, G.~Hinton, A.~Krizhevsky, I.~Sutskever, and R.~Salakhutdinov.
\newblock Dropout: A simple way to prevent neural networks from overfitting.
\newblock {\em J. Mach. Learn. Res.}, 15(1):1929--1958, 2014.

\bibitem{Tola08}
E.~Tola, V.Lepetit, and P.~Fua.
\newblock {A Fast Local Descriptor for Dense Matching}.
\newblock In {\em CVPR}, 2008.

\bibitem{Lebedev2014}
L.~Vadim, G.~Yaroslav, R.~Maksim, O.~Ivan, and L.~Victor.
\newblock Speeding-up convolutional neural networks using fine-tuned
  cp-decomposition.
\newblock {\em CoRR}, abs/1412.6553, 2014.

\bibitem{Vanhoucke2011}
V.~Vanhoucke, A.~Senior, and M.~Z. Mao.
\newblock Improving the speed of neural networks on cpus.
\newblock In {\em Deep Learning and Unsupervised Feature Learning Workshop,
  NIPS 2011}, 2011.

\bibitem{vedaldi15matconvnet}
A.~Vedaldi and K.~Lenc.
\newblock Matconvnet -- convolutional neural networks for matlab.
\newblock In {\em Proceeding of the {ACM} Int. Conf. on Multimedia}, 2015.

\bibitem{Viola01}
P.~Viola and M.~Jones.
\newblock Rapid object detection using a boosted cascade of simple features.
\newblock In {\em CVPR}, 2001.

\bibitem{Caffe2014}
J.~Yangqing, S.~Evan, D.~Jeff, K.~Sergey, L.~Jonathan, G.~Ross, G.~Sergio, and
  D.~Trevor.
\newblock Caffe: Convolutional architecture for fast feature embedding.
\newblock In {\em Proceedings of the ACM International Conference on
  Multimedia}, 2014.

\end{thebibliography}
}

\section{Appendix}
\label{App:LemmaProof}

Here we prove the statements of lemma~\ref{lemma:L1Delta} from Section~\ref{sub:TrainDetails}.

\begin{proof}

\begin{enumerate}
\item

From the definition $\bar{\delta}^{(b,1)} = \delta^{(b,1)} - \frac{\#\delta{(b,1)}}{\#\delta^b} \delta^b$, so $\delta^{(b,1)} = \frac{\#\delta^{(b,1)}}{\#\delta^b}\delta^b + \bar{\delta}^{(b,1)}$. Also, since $\delta^b= \delta^{(b,0)} + \delta^{(b,1)}$, we have

\begin{eqnarray*}
\delta^{(b,0)} &= &\delta^b-\delta^{(b,1)}  \\
 &=& \delta^b - ( \frac{\#\delta^{(b,1)}}{\#\delta^b}\delta^b + \bar{\delta}^{(b,1)} ) \\
 &=& \delta^b(1-\frac{\#\delta^{(b,1)}}{\#\delta^b}) - \bar{\delta}^{(b,1)} \\
 &=& \delta^b \frac{\#\delta^{(b,0)}}{\#\delta^b} - \bar{\delta}^{(b,1)}
\end{eqnarray*}

Hence, for weights $w_0,w_1$, 
\begin{eqnarray*}
w_0 \delta^{(b,0)} &+& w_1 \delta^{(b,1)}  \\
&=& w_0( \frac{\#\delta^{(b,0)}}{\#\delta^b}\delta^b - \bar{\delta}^{(b,1)} ) \\
& + & w_1( \frac{\#\delta^{(b,1)}}{\#\delta^b}\delta^b + \bar{\delta}^{(b,1)} ) \\
&=& (w_0+w_1)\delta^b + (w_1-w_0)\bar{\delta}^{(b,1)}
\end{eqnarray*}

So $w_b=(w_0+w_1)$ and $w_\Delta = w_1-w_0$ fulfill the lemma's statement.

\item 

\begin{eqnarray*}
R^c(\bar{\delta}^{(b,1)}) & = & | \sum_{i,p} g^c_i \bar{\delta}^{(b,1)}_{i,p} | \\
&=& | \sum_{i,p} g^c_i \delta^{(b,1)}_{i,p} -  \frac{\#\delta^{(b,1)}}{\#\delta^b} \sum_{i,p} g^c_i \delta^b_{i,p} |
\end{eqnarray*}

Using $\delta^{(b,0)} = \delta^b- \delta^{(b,1)}$ we continue to
\begin{eqnarray*}
& = &| \sum_{i,p} g^c_i(\delta^b_{i,p} -\delta^{(b,0)}_{i,p}) - \frac{\#\delta^b - \#\delta^{(b,0)}}{ \#\delta^b}\sum_{i,p} g^c_i \delta^b_{i,p} | \\
& = &| -\sum_{i,p} g^c_i \delta^{(b,0)}_{i,p} + \frac{\#\delta^{(b,0)}}{\#\delta^b} \sum_{i,p} g^c_i \delta^b_{i,p} | \\
& = & R^c(\bar{\delta}^{(b,0)})
\end{eqnarray*}

\item
\begin{eqnarray*}
 && | \sum_{i,p}  (g^c_i  - E[g^c_i|b])  \delta^{(b,1)}_{i,p}| \\
&=&  | \sum_{i,p} (g^c_i  - \frac{\sum_{i',p'} g^c_{i'}\delta^b_{i',p'}}{\#\delta^b} )  \delta^{(b,1)}_{i,p}| \\
 &=& |\sum_{i,p} g^c_i \delta^{(b,1)}_{i,p} - \frac{\#\delta^{(b,1)}}{\#\delta^b} \sum_{i,p} g^c_i \delta^b_{i,p} | \\
&=& R^c(\bar{\delta}^{(b,1)}) 
\end{eqnarray*}
\end{enumerate}
\end{proof}

\end{document}